\documentclass[sigconf]{acmart}
\usepackage{algorithm}
\usepackage{algorithmic}
\usepackage{multirow}
\usepackage{float}
\definecolor{MyRed}{HTML}{BE0712}
\definecolor{MyBlue}{HTML}{2D729C}
\def\ie{{\em i.e.}}
\def\eg{{\em e.g.}}

\newcommand{\heading}[1]{\noindent\textbf{#1}}

\AtBeginDocument{%
  \providecommand\BibTeX{{%
    \normalfont B\kern-0.5em{\scshape i\kern-0.25em b}\kern-0.8em\TeX}}}

\copyrightyear{2021}
\acmYear{2021}
\setcopyright{acmcopyright}\acmConference[MM '21]{Proceedings of the 29th ACM International Conference on Multimedia}{October 20--24, 2021}{Virtual Event, China} \acmBooktitle{Proceedings of the 29th ACM International Conference on Multimedia (MM '21), October 20--24, 2021, Virtual Event, China}
\acmPrice{15.00}
\acmDOI{10.1145/3474085.3475200}
\acmISBN{978-1-4503-8651-7/21/10}

\acmSubmissionID{237}

\begin{document}
\fancyhead{} 
\title{Towards Cross-Granularity Few-Shot Learning: Coarse-to-Fine Pseudo-Labeling with Visual-Semantic Meta-Embedding}

\author{Jinhai Yang, Hua Yang$^{*}$, Lin Chen}
	\affiliation{\institution{Institute of Image Communication and Network Engineering, Shanghai Jiao Tong University, China}}
    \affiliation{\institution{Shanghai Key Lab of Digital Media Processing and Transmission, Shanghai, China}}
    \affiliation{\institution{MoE Key Lab of Artificial Intelligence, AI Institute, Shanghai Jiao Tong University, China}}
	\thanks{$^*$ Hua Yang is the corresponding author (E-mail: hyang@sjtu.edu.cn).}

\renewcommand{\shortauthors}{Jinhai Yang, Hua Yang, and Lin Chen}

\begin{abstract}
Few-shot learning aims at rapidly adapting to novel categories with only a handful of samples at test time, which has been predominantly tackled with the idea of meta-learning.
However, meta-learning approaches essentially learn across a variety of few-shot tasks and thus still require large-scale training data with fine-grained supervision to derive a generalized model, thereby involving prohibitive annotation cost.
In this paper, we advance the few-shot classification paradigm towards a more challenging scenario, \ie, cross-granularity few-shot classification, where the model observes only coarse labels during training while is expected to perform fine-grained classification during testing.
This task largely relieves the annotation cost since fine-grained labeling usually requires strong domain-specific expertise.
To bridge the cross-granularity gap, we approximate the fine-grained data distribution by greedy clustering of each coarse-class into pseudo-fine-classes according to the similarity of image embeddings.
We then propose a meta-embedder that jointly optimizes the visual- and semantic-discrimination, in both instance-wise and coarse class-wise, to obtain a good feature space for this coarse-to-fine pseudo-labeling process.
Extensive experiments and ablation studies are conducted to demonstrate the effectiveness and robustness of our approach on three representative datasets.
\end{abstract}




\begin{teaserfigure}
\includegraphics[width=\linewidth]{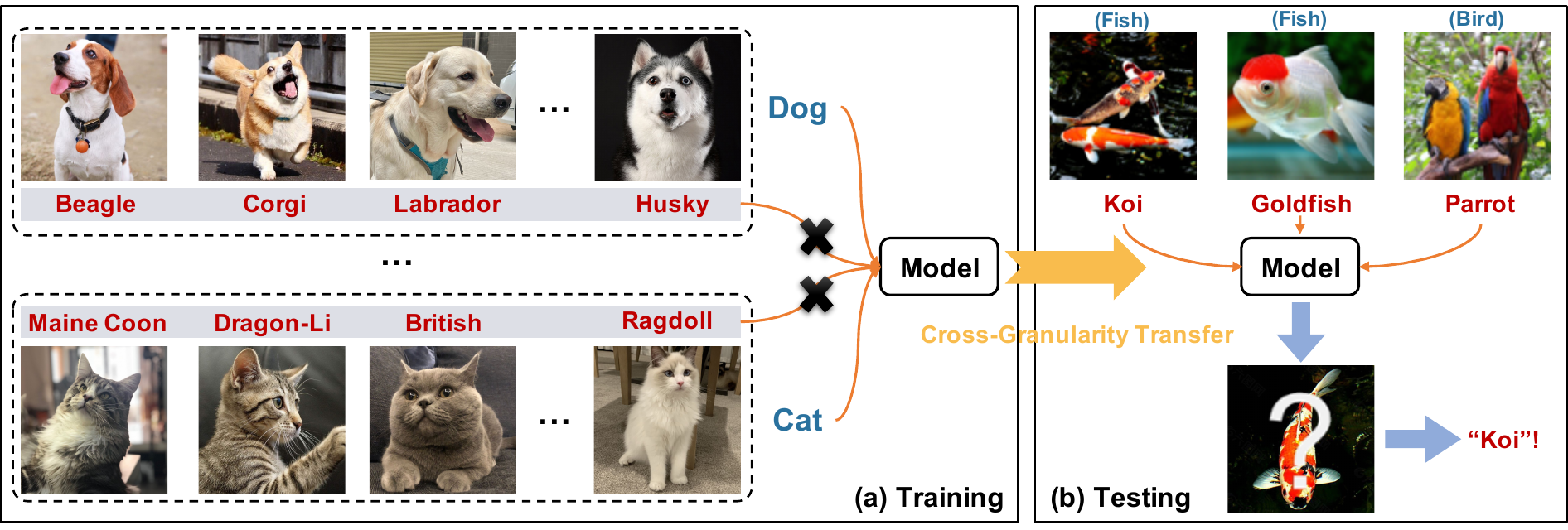}
  \caption{Illustrating the cross-granularity few-shot image classification problem (along with a 3-way 1-shot testing case).
  (a) 
  Training with only \textit{coarse} class labels~(in \textcolor{MyBlue}{blue}).
  (b)
  Testing on \textit{novel} \textit{few-shot} tasks with \textit{unseen} \textit{fine} class labels~(in \textcolor{MyRed}{red}).
}
\label{fig:task}
\end{teaserfigure}

\maketitle
\begin{figure*}[ht]
\includegraphics[width=\linewidth]{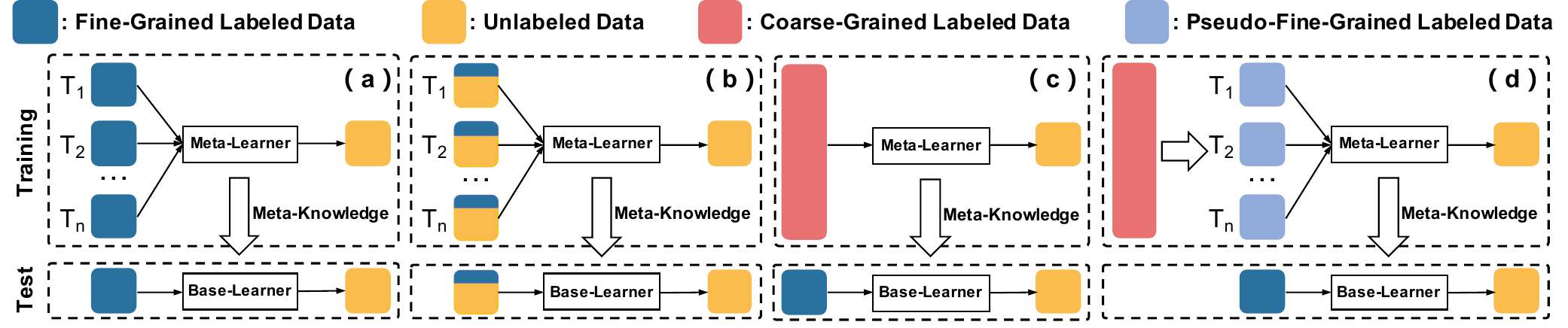}
  \caption{
  Comparisons between the proposed learning framework and existing meta-learning paradigms.
  (a) Fully-supervised meta-learning, the most regular setting, where all training samples are associated with fine labels (referred to as \textit{fine-to-fine} in this paper).
  (b) Semi-supervised meta-learning. 
  A pool of unlabeled samples is introduced as auxiliary data.
  (c) Inexactly-supervised meta-learning (ours). 
  Training samples are associated with only coarse labels.
  (d) Our core idea to tackle this new problem. 
  We propose to approximate the fine-grained task distribution from coarse classes for cross-granularity learning.
}
\label{fig:problem}
\end{figure*}

\vspace{-0.1in}
    {\fontsize{8pt}{8pt} \selectfont\textbf{ACM Reference Format:}\\Jinhai Yang, Hua Yang, and Lin Chen. 2021. Towards Cross-Granularity Few-Shot Learning: Coarse-to-Fine Pseudo-Labeling with Visual-Semantic Meta-Embedding. In \textit{Proceedings of the 29th ACM International Conference on Multimedia (MM ’21), October 20–24, 2021, Virtual Event, China.} ACM, New York, NY, USA, 10 pages. https://doi.org/10.1145/3474085.3475200}
	\vspace{0.1in}

\section{Introduction}
As a hallmark of intelligence, humans can easily learn new concepts with scarce samples. 
In stark contrast, recent advances in deep learning models usually demand immense quantities of data with fine-grained annotations to learn robust systems~\cite{krizhevsky2012imagenet,he2016deep,attention}. 
This requirement severely limits their practicality due to \textbf{two issues}: 
\textbf{(1) the difficulty of collecting massive training samples} and 
\textbf{(2) the high cost of exhaustively fine-grained data labeling}.  
To address the first issue, there has been growing interest in few-shot learning~(FSL)~\cite{koch2015siamese,vinyals2016matching,snell2017prototypical,sung2018learning,tian2020rethink,du2021fewshot}, where a model learns from a set of base classes for the adaptability to unseen classes with only a few samples.
A testing episode consists of a labeled (few-shot) support set for task-specific learning and a query set for evaluating the acquired model.
Recent years have witnessed the success of meta-learning~(ML)~\cite{schmidhuber1987evolutionary,thrun1998learning,finn2017model,sun2019meta,jamal2019task} on FSL and many downstream problems~\cite{zheng2020learning,li2020meta,fu2020depth,cao2020task}, partially attributed to the episodic training strategy that mimics the low-data regime at training time and acquires transferable knowledge from a variety of analogous episodes. 
As shown in Fig.~\ref{fig:problem}~(a), \textbf{most ML algorithms are fully supervised and thus still rely heavily on fine-grained annotations.} 
The second issue remains a stumbling block.

In this paper, we consider a more practical scenario, \ie, cross-granularity few-shot classification (CGFSC), to alleviate the annotation burden.
As depicted in Fig.~\ref{fig:task}, whereas the model is trained with only coarse class labels (\eg, animals) rather than the \textbf{exact} fine labels (\eg, breeds), it is required to generalize well on unseen fine classes.
This objective discrepancy makes it a challenging task since the model must learn the internal structure of a given coarse class for cross-granularity learning.
As illustrated in Fig.~\ref{fig:problem}, different from existing fully-supervised (Fig.~\ref{fig:problem}~(a)) or semi-supervised (Fig.~\ref{fig:problem}~(b)) paradigms that meta-learn in a \textbf{fine-to-fine} (F2F) manner, we cast the CGFSC as an inexactly-supervised meta-learning problem (Fig.~\ref{fig:problem}~(c)) that meta-learns in a \textbf{coarse-to-fine} (C2F) manner.
\textbf{Inexact supervision} is a typical type of weak supervision, where the training data are only associated with coarse-grained labels~\cite{zhou2018brief}.
Although \cite{liu2019prototype} has proposed to utilize coarse labels in FSL, they use coarse labels as additional information (to fine labels) to boost the performance, while CGFSC deals with coarse labels as the only guidance.
As indicated in Tab.~\ref{tab:dataset}, the number of coarse classes is far smaller than that of fine classes (\eg, 34 vs 608 for tieredImageNet). 
The coarse labels are easier to produce in real-world applications since fine-grained annotation usually demands sophisticated domain-specific knowledge.

To tackle this new problem, we propose a Visual-Semantic Meta-Embedder~(VSME) with a Coarse-to-Fine~(C2F) pseudo-labeling algorithm to generate the approximate fine-grained training classes.
Unlike the pseudo-labeling technique widely used in the field of semi-supervised learning~\cite{lee2013pseudo,shi2018transductive,iscen2019label,xie2020self}, our C2F pseudo-labeling produces fine labels for coarse-labeled data.
We cluster each coarse class into pseudo-fine-classes via greedy similarity matching to establish a task distribution of \textbf{analogous granularity} to the target testing tasks, as shown in Fig.~\ref{fig:problem} (d). 
Recent works have shown that the fast adaptation of ML crucially depends on feature reuse~\cite{Raghu2020Rapid}, and a good feature embedding trained on the merged base-classes performs favorably in the standard ML setting~\cite{tian2020rethink,chen2018a,Dhillon2020A,wang2020cooperative}.
Therefore, we present the VSME to learn from all training samples and then generate image embeddings for the similarity matching, along with a Base-Embedder as the \textbf{downstream} few-shot learner~(\ie, both the meta- and base-learner of the FSL stage in Fig.~\ref{fig:problem}~(d)).

\begin{figure}[t]
\includegraphics[width=\linewidth]{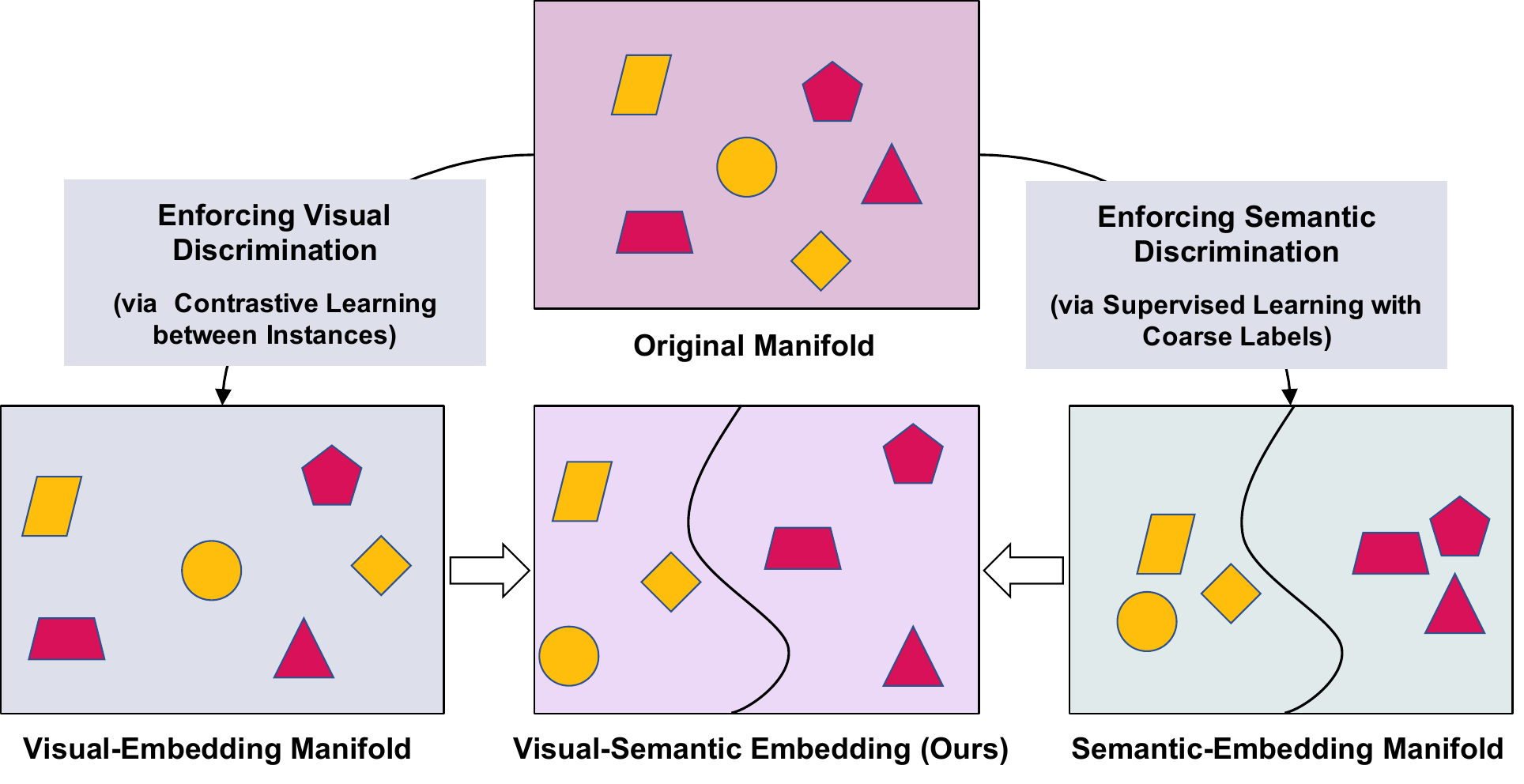}
  \caption{The motivation of optimizing visual- and semantic-discrimination jointly for the Meta-Embedder. 
  Colors represent different \textit{coarse} classes and shapes represent different \textit{fine} classes. 
  Supervised learning with coarse labels promotes semantic discrimination between coarse classes and instance-wise contrastive learning helps explore the internal structure of the coarse class manifold.
}
\label{fig:motivation}
\end{figure}

We jointly optimize the visual- and semantic-discrimination between the VSME features, as illustrated in Fig.~\ref{fig:motivation}.
Instance-based embedding methods~\cite{wu2018unsupervised,dosovitskiy2015discriminative} take each instance as an individual class.
\cite{ye2019unsupervised} utilizes \textbf{visual transformations} to augment the samples and then conducts contrastive learning within mini-batches. 
It focuses on the visual side and helps \textbf{exploit the internal structure} of each coarse class.
Supervised embedding~\cite{oh2016deep,oh2017deep,harwood2017smart,tian2020rethink} explicitly maximizes inter-class variation while minimizes intra-class variation.
These methods provide more \textbf{semantic information} and help \textbf{explore the external relation} between coarse classes, but also make the fine-grained classes inside one coarse class inseparable unfavorably.
In this paper, we combine the \textbf{complementary} strengths of instance-wise supervision and coarse-class labels to integrate both visual- and semantic-discrimination.

\textbf{Our contributions are summarized as follows}:
\textbf{(1)}
We propose the cross-granularity few-shot classification task and cast it as the inexactly-supervised meta-learning problem, which learns from coarse labels for few-shot testing on unseen fine classes.
\textbf{(2)}
We devise a Visual-Semantic Meta-Embedder to encode images and the Coarse-to-Fine pseudo-labeling that performs similarity matching with the embeddings to generate pseudo-fine-classes.
\textbf{(3)}
We conduct extensive experiments and analysis to show the effectiveness and robustness of the proposed approach on three datasets.

\section{Related Works}
\noindent \textbf{Few-shot learning.}
FSL desires the model capability of adapting or generalizing well to unseen tasks with little data. 
Meta-learning~(ML)~\cite{schmidhuber1987evolutionary}, also referred to as \textit{learning to learn}~\cite{thrun1998learning}, has become a popular way to tackle the FSL problem in recent years.
ML is usually accompanied by the episodic training strategy~\cite{vinyals2016matching} to learn high-level transferable knowledge over a sampled distribution of analogous few-way few-shot tasks~(typically 5-way 1/5-shot), mimicking the data limitation at test time. 
In general, there are three types of episodic ML methods: 
\textbf{(1) model-based:} building a tailored model specially designed for fast learning via,~\eg, an external memory storage~\cite{santoro2016meta}, an independent meta-learner~\cite{munkhdalai2017meta}, or a neural attentive learner~\cite{mishra2018simple}; 
\textbf{(2) metric-based:} learning components of a differentiable weighted $k$-nearest neighbors~($k$-NN) classifier, such as prototypical representations~\cite{snell2017prototypical}, a relation measurement~\cite{sung2018learning} or a task-dependent adaptive metric~\cite{tadam2018_66808e32}; 
\textbf{(3) optimization-based:} learning an efficient LSTM~\cite{hochreiter1997long} optimizer~\cite{Ravi2017OptimizationAA}, an optimal model initialization~\cite{finn2017model}, a good feature space with convex base-learners, {\em etc.} 
Most recently, many works~\cite{tian2020rethink,wang2020cooperative,chen2020new,Dhillon2020A} have noted that simply training an embedder on all base classes without episodic sampling performs no worse than other complicated ML methods.
In this work, we meta-learn a feature space with VSME to specify the fine class distribution for a base-embedder~(BE) to learn from.
We find that episodic training further improves the performance of BE in the CGFSC task, after pre-training on the pseudo-classes.

\heading{Deep embedding learning.}
Different from general feature learning, deep embedding learning seeks to make the distance between the learned features consistent with the visual similarity or semantic relation of the input images. 
\textbf{Supervised} methods~\cite{oh2016deep,oh2017deep,harwood2017smart,tian2020rethink} tend to learn a feature space by directly increasing inter- while decreasing intra-class variation. 
Note that with coarse labels, these methods would essentially \textit{eliminate} the internal separation within the coarse class manifold, making it hard to distinguish between the contained fine-grained classes. 
Instead, existing \textbf{unsupervised} methods take advantage of either self-supervised learning or instance-wise contrastive learning.
\textbf{Self-supervised} methods create proxy tasks~\cite{doersch2015unsupervised,zhang2016colorful,pathak2016context} to drive feature learning without any external guidance, while \textbf{instance-wise contrastive} methods treat each sample as a different class~\cite{wu2018unsupervised,dosovitskiy2015discriminative}, and are usually built on the intuition that different augmentations of an image should be recognized as the same instance~\cite{ye2019unsupervised,chen2020simple,he2020momentum}.  
As shown in Fig.~\ref{fig:motivation}, we combine the strength of supervised~(with coarse label) and instance-based methods to work in synergy.

\noindent \textbf{Meta-learning with limited supervision.}
Although aiming at few-shot testing, most ML methods still heavily depend on large-scale fine-grained annotations. 
Recently, some studies have begun to explore meta-learning with different types of limited annotations. 
\cite{ren2018meta} pioneered \textbf{semi-supervised} meta-learning, which mixes a small labeled support set with a large pool of unlabeled data, and proposed an extension of Prototypical Networks~\cite{snell2017prototypical} to better utilize the unlabeled set. 
Under the same setting,~\cite{li2019learning} introduced a self-training strategy and a cherry-picking network to select and annotate the unlabeled set. 
\textbf{Unsupervised} meta-learning purely learns from unsupervised data, but, meanwhile suffers dramatic performance drops due to the lack of annotations.
Existing works attempted to synthesize training tasks by clustering embeddings~\cite{hsu2018unsupervised} learned unsupervised~\cite{caron2018deep,DBLP:conf/iclr/DonahueKD17} or random sampling and augmentation~\cite{Khodadadeh2019unsupervised}.
In our framework, we use only coarse labels to provide moderate supervision to balance the \textit{trade-off} between annotation overhead and performance.  

\noindent \textbf{Learning with inexact supervision.}
Since the supervision information given by coarse labels is not as exact as desired, it is well-known as \textit{inexact supervision}, one of the most common types of weak supervision~\cite{zhou2018brief}. 
Some works~\cite{ristin2015categories,guo2018cnn} have considered learning problems entailing the mixture of coarse and fine labels, training the model with either balanced~\cite{taherkhani2019weakly} or unbalanced~\cite{robinson2020strength} batches containing both coarse- and fine-labeled data. 
In the context of FSL, there are few works concentrated on coarse-grained information.
\cite{liu2019prototype} proposed a prototype propagation mechanism to effectively exploit the hierarchical labels of ImageNet~\cite{deng2009imagenet} and learn a multi-level directed acyclic prototype graph. 
However, this method treats coarse labels as auxiliary information for few-shot classification tasks and assumes that the class hierarchy is known in advance~(even during testing).
Differently, we observe only coarse labels of base classes and the target classes belong to unseen coarse classes.
\textbf{To our best knowledge, we are the first study to leverage pure inexact supervision in FSL.}

\begin{figure*}[t]
\includegraphics[width=\linewidth]{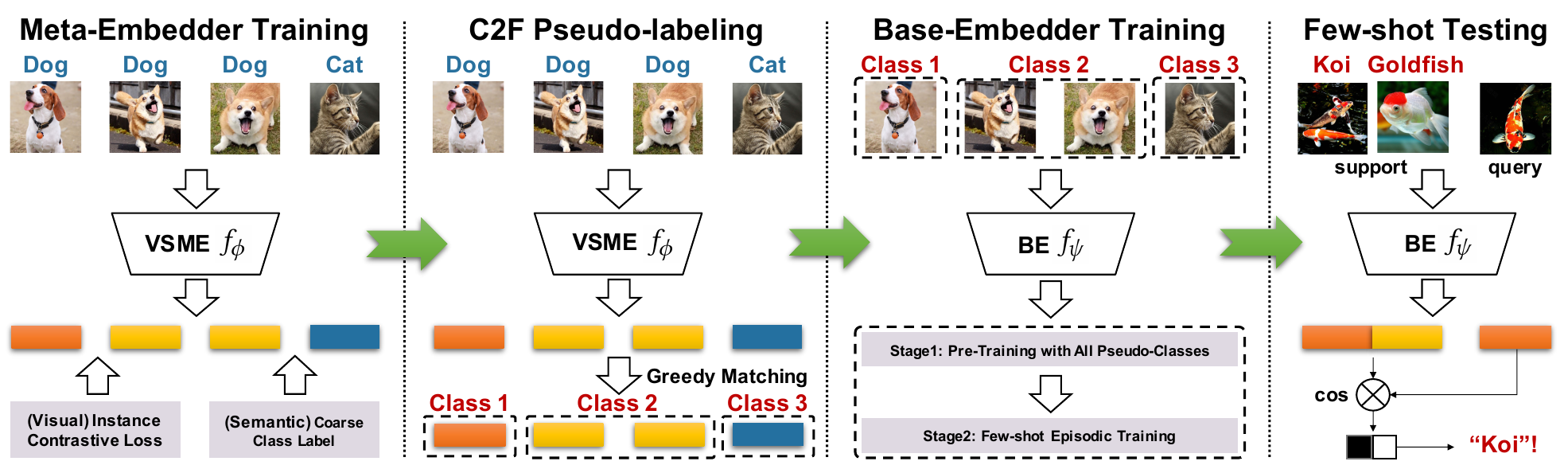}
  \caption{
Overview of our framework.
VSME: Visual-Semantic Meta-Embedder.
BE: Base-Embedder.
\textcolor{MyBlue}{Blue}: coarse.
\textcolor{MyRed}{Red}: fine.
}
\label{fig:method}
\end{figure*}
\section{Problem Formulation}
In \textbf{standard few-shot classification}, 
given a meta-training set $\mathcal{D}_{train}$ containing samples of \textit{base classes} $\mathcal{C}_{base}^{fine}$, the goal is to achieve high accuracy on classification episodes sampled from a testing set $\mathcal{D}_{test}$ that consists of \textit{novel classes} $\mathcal{C}_{novel}^{fine}$.
The novel classes are totally different from the base classes, \ie, $\mathcal{C}_{base}^{fine}\cap\mathcal{C}_{novel}^{fine}=\emptyset$.
An $N$-way $K$-shot episode is defined as a tuple $(\mathcal{S}_{support},\mathcal{S}_{query})$, where the \textit{support set} $\mathcal{S}_{support}$ contains $N$ classes with $K$ samples in each class and the \textit{query set} $\mathcal{S}_{query}$ contains the same $N$ classes with $Q$ unlabeled samples in each class.
The episodic objective is to correctly classify the $N\times Q$ query samples into the $N$ classes, taking $\mathcal{S}_{support}$ as a reference.

\noindent In \textbf{cross-granularity few-shot classification}, the training set $\mathcal{D}_{train}=\{(\mathbf{x}_i,\tilde{y}_i)\}$ includes pairs of images $x_i \in \mathcal{X}$ and coarse labels $\tilde{y}_i \in \mathcal{C}_{base}^{coarse}$ rather than the fine label $y_i\in \mathcal{C}_{base}^{fine}$.
Taking ImageNet~\cite{deng2009imagenet} as an example, instead of being labeled exactly, the images of class ``lamp'' and ``bookcase'' may share the coarse-category label ``furnishing''.  
Besides the fine labels, the coarse labels between base classes and novel classes should also be \textbf{non-overlapping}, that is, $\mathcal{C}_{base}^{coarse}\cap\mathcal{C}_{novel}^{coarse}=\emptyset$, making CGFSC more challenging than standard FSL.
Notably, in CGFSC the goal is still to minimize the few-shot test error on the \textbf{fine} \textbf{novel} classes $\mathcal{C}_{novel}^{fine}$.

\section{Method}
\heading{Framework overview.}
An overview of the proposed framework is illustrated in Fig.~\ref{fig:method}.
The training process consists of three procedures:
\textbf{(1) Meta-Embedder Training}:
Train a Convolutional Neural Network (CNN) as the meta-embedding function $f_\phi$ with joint instance-wise supervision and coarse class-wise supervision on the meta-training set $\mathcal{D}_{train}$. 
\textbf{(2) C2F Pseudo-Labeling}:
Use $f_\phi$ to derive image embeddings for all samples in $\mathcal{D}_{train}$ and then conduct similarity matching to pseudo-label each coarse class $\mathcal{C}_{base}^{coarse}$ into pseudo-fine-classes $\mathcal{C}_{base}^{pseudo}$, according to the similarity of the corresponding features. 
\textbf{(3) Base-Embedder Training}: 
Train another CNN as the base-embedding function $f_\psi$ on the pseudo-dataset $\mathcal{D}_{train}^{pseudo}$ to excavate transferable knowledge.
Last, for the \textbf{few-shot testing}, we take the fixed $f_\psi$ as a feature extractor and utilize a nearest-prototype classifier with cosine similarity to predict the fine label for $\mathcal{S}_{query}$ using a fine-labeled $\mathcal{S}_{support}$.

\subsection{Visual-Semantic Meta-Embedder}
\begin{figure*}[t]
\includegraphics[width=\linewidth]{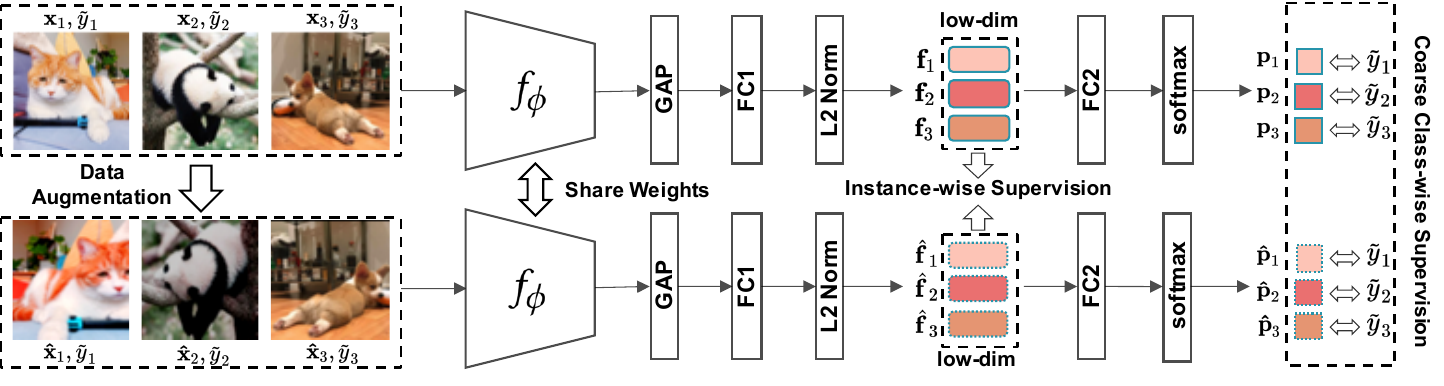}
  \caption{
  Illustration of the Visual-Semantic Meta-Embedder.
  Original and augmented images (via visual transformations) fed into bilateral branches in a Siamese structure.
  The instance-wise supervision keeps the pre- and post-augmentation features of an instance \textit{invariant} while the features of different instances \textit{separated}, thus enforcing visual discrimination between the low-dimensional embeddings.
  Meanwhile, the embeddings are further fed into a fully-connected classifier.
  The coarse class-wise supervision enforces the semantic discrimination via the co-training of coarse-level classification.
  }
\label{fig:embedding}
\end{figure*}

In this section, we elaborate on the details of the proposed VSME, which approximates both visual discrimination and semantic discrimination in the latent space. 
As shown in Fig.~\ref{fig:embedding}, VSME simultaneously exploits two types of supervision: (1) instance-wise supervision and (2) coarse class-wise supervision.
Although they share the same objective of pulling similar images together and pushing dissimilar ones away in a compact space, they encourage feature discrimination on quite different aspects. 
The instance-wise supervision imposes consistency regularization~\cite{mix} to ensure the model produces identical output distribution when its inputs are visually perturbed, hence it focuses more on the \textbf{visual} side.
Differently, the coarse class-wise supervision enforces the feature vectors discriminative in the \textbf{semantic} aspect.

We use four methods (RandomResizedCrop, ColorJitter,
RandomGrayscale, RandomHorizontalFlip) to augment the images.
For a coarsely-labeled sample $\left(\mathbf{x},\tilde{y}\right)\in \mathcal{D}_{train}$, let~$\hat{\mathbf{x}}$ denote the augmented sample. 
For simplicity, let $\mathbf{f}$ and $\hat{\mathbf{f}}$ denote their embedding features. 
As shown in Fig.~\ref{fig:embedding}, an image $\mathbf{x}$ is first encoded by $f_\phi$, next passed through the global average pooling (GAP) to take the average of each feature map, and then converted to a low-dimensional embedding $\mathbf{f}\in\mathbb{R}^{M}$ with a fully-connected layer FC1 (assuming the output dimension is $M$).
The embeddings are $\ell_{2}$ normalized, thus $\left\|\mathbf{f}\right\|_{2}=1$, therefore the cosine similarity can be efficiently calculated by the dot product.
Following~\cite{wu2018unsupervised,ye2019unsupervised}, we use the softmax function with cosine similarity on top of the embeddings to match the instances.
For the $i$-th sample $\mathbf{x}_i$ in a batch $\mathcal{B}$, the probability of classifying $\hat{\mathbf{x}}_{i}$ as the original instance $\mathbf{x}_i$ is defined by
\begin{equation}
    P\left(\mathbf{x}_i | \hat{\mathbf{x}}_{i}\right)=\frac{\exp \left(\mathbf{f}_{i}^{T} \hat{\mathbf{f}}_{i} / \tau\right)}{\sum_{\mathbf{x}_{k}\in \mathcal{B}} \exp \left(\mathbf{f}_{k}^{T} \hat{\mathbf{f}}_{i} / \tau\right)}~,
\end{equation}
where $\tau$ is the temperature \cite{goodfellow2016deep,guo2017calibration} scaling the entropy of the output probability distribution.
The probability of another instance $x_{j}$ being falsely recognized
as the $i$-th instance is
\begin{equation}
    P\left(\mathbf{x}_i | \mathbf{x}_{j}\right)=\frac{\exp \left(\mathbf{f}_{i}^{T} \mathbf{f}_{j} / \tau\right)}{\sum_{\mathbf{x}_k\in\mathcal{B}} \exp \left(\mathbf{f}_{k}^{T} \mathbf{f}_{j} / \tau\right)}, j \neq i~.
\end{equation}
The \textbf{visual-side} negative log-likelihood loss within the batch is
\begin{equation}
    \mathcal{L}_{vis}=-\sum_{\mathbf{x}_i\in\mathcal{B}} \log P\left(\mathbf{x}_i | \hat{\mathbf{x}}_{i}\right)-\sum_{\mathbf{x}_i\in\mathcal{B}} \sum_{\mathbf{x}_j\neq\mathbf{x}_i} \log \left(1-P\left(\mathbf{x}_i | \mathbf{x}_{j}\right)\right)~.
\end{equation}
 
To harness the coarse labels $\tilde{y}\in \mathcal{C}_{base}^{coarse}=\{1,2,\ldots,C\}$ for the semantic discrimination property, the embedding features $\mathbf{f}$ are sent to a linear classifier (FC2) $\mathbf{W}=[\mathbf{W}_1,\mathbf{W}_2,\cdots,\mathbf{W}_C] \in \mathbb{R}^{M \times C}$ with a softmax layer for classification of the coarse classes. 
The probability of $\mathbf{x}_{i}$ belonging to class $c$ is defined by
\begin{equation}
    P\left(c | \mathbf{x}_{i}\right)=\frac{\exp \left(\mathbf{W}_{c}^{T} \mathbf{f}_{i}\right)}{\sum_{k=1}^{C} \exp \left(\mathbf{W}_{k}^{T} \mathbf{f}_{i} \right)}~.
\end{equation}
On the other hand, the augmented image $\hat{\mathbf{x}}_i$ should be classified into the same class as that of the original one. 
The probability of the augmented image $\hat{\mathbf{x}}_{i}$ being classified to class $c$ is denoted by
\begin{equation}
    P\left(c | \hat{\mathbf{x}}_{i}\right)=\frac{\exp \left(\mathbf{W}_{c}^{T} \hat{\mathbf{f}}_{i}\right)}{\sum_{k=1}^{C} \exp \left(\mathbf{W}_{k}^{T} \hat{\mathbf{f}}_{i} \right)}~.
\end{equation}
Let $\tilde{y}_{i,j}$ denote the j-th class label of instance $\mathbf{x}_{i}$. 
The \textbf{semantic-side} negative log-likelihood loss within the batch is
\begin{equation}
    \mathcal{L}_{sem}=-\sum_{\mathbf{x}_i\in\mathcal{B}} \sum_{j=1}^{C}\tilde{y}_{i,j} \log \left[P\left(j |\mathbf{x}_{i}\right)P\left(j |\mathbf{\hat{x}}_{i}\right)\right]~.
\end{equation} 
The joint loss of the VSME is the sum of the two loss items,
\begin{equation}
    \mathcal{L}_{meta}(\mathcal{B};\phi)=\mathcal{L}_{vis}+  \mathcal{L}_{sem}~,
\end{equation}
and the optimal parameters are $
    \phi^{*}=\underset{\phi}{\arg \min }~ \mathbb{E}_{\mathcal{B}}\left[\mathcal{L}_{meta}\left(\mathcal{B};\phi\right)\right].$
\subsection{Coarse-to-Fine Pseudo-Labeling}
Unlike the pseudo-labeling in the usual sense~\cite{lee2013pseudo,li2019learning}, we propose the coarse-to-fine pseudo-labeling that takes as input an image with a coarse label $(\mathbf{x},\tilde{y})\in\mathcal{D}_{train}$ and assigns a fine-grained pseudo-label to it. 
As outlined in Alg.~\ref{alg:1}, this process continually selects random images as the root samples and then makes locally-optimal choices to produce the pseudo-sub-categories, until the pseudo-labeled training dataset $\mathcal{D}_{train}^{pseudo}$ is entirely derived. 
In this process, we still conduct image matching with cosine similarity (dot product) on top of the corresponding features encoded by the fixed $f_{\phi^{*}}$. 

\begin{algorithm}[tb]
\caption{Coarse-to-Fine (C2F) Pseudo-Labeling}
\begin{algorithmic}[1] 
\REQUIRE $N_{s}$: Number of samples per pseudo-fine-class
\REQUIRE $C$: Number of coarse classes, $C=\left|\mathcal{C}_{base}^{coarse}\right|$
\REQUIRE $\mathcal{D}_{train}=\left\{\left(\mathbf{x}_{i}, \tilde{y}_{i}\right)\right\}$: Training data with coarse labels
\STATE Initialize the pseudo-dataset $\mathcal{D}_{train}^{pseudo}=\left\{\right\}$ 
\FOR{class $c$ from 1 to $C$}
\STATE Retrieval $\widetilde{\mathcal{D}}_{c}=\left\{\mathbf{x}_{i}|\tilde{y}_{i}=c,(\mathbf{x}_{i},\tilde{y}_{i})\in \mathcal{D}_{train}\right\}$, let $N=|\widetilde{\mathcal{D}}_{c}|$
\STATE Get the embeddings $\mathbf{F}_{c}\in \mathbb{R}^{M\times N}$ of the samples in $\widetilde{\mathcal{D}}_{c}$\label{tag:similarity}
\STATE Obtain the similarity matrix $\mathbf{S}=\mathbf{F}_{c}^{T}\mathbf{F}_{c}\in \mathbb{R}^{N\times N}$
\WHILE{number of the remaining samples of $\widetilde{\mathcal{D}}_{c}\geq N_{s}$}
\STATE Randomly select $\mathbf{x}_{j}$ from $\widetilde{\mathcal{D}}_{c}$ as the root of new class $c_{n}$
\STATE Retrieval $\{\mathbf{x}_{k}\}_{k=1}^{N_{s}-1}$ with top similarity to $\mathbf{x}_{j}$
\STATE $\mathcal{D}_{train}^{pseudo}\leftarrow\mathcal{D}_{train}^{pseudo}\cup\left\{\left(\mathbf{x}_{j},c_{n}\right)\right\}\cup\{(\mathbf{x}_{k},c_{n})\}_{k=1}^{N_{s}-1}$
\STATE Remove the selected $N_{s}$ samples from $\widetilde{\mathcal{D}}_{c}$
\ENDWHILE
\STATE Discard the rest of the samples of $\widetilde{\mathcal{D}}_{c}$
\ENDFOR
\end{algorithmic}
 \label{alg:1}
\end{algorithm}

\subsection{Base-Embedder}\label{sec:be}
Once all coarse classes in $\mathcal{D}_{train}$ are divided into fine-grained sub-categories via C2F pseudo-labeling, we train another embedding model $f_\psi$ as the few-shot learner with the pseudo-dataset $\mathcal{D}_{train}^{pseudo}$. 
As shown in Fig.~\ref{fig:method}, the training of $f_\psi$ takes place in two consecutive stages: (1) pre-training with all pseudo-fine-classes and (2) episodic training on few-way few-shot episodes sampled from $\mathcal{D}_{train}^{pseudo}$.

In the \textbf{pre-training stage}, our goal is to learn a set of parameters $\theta$ as the initialization of $f_\psi$ via a classification task on the complete pseudo-dataset.
Let $\mathcal{L}_{ce}$ denotes the cross-entropy loss, the optimization objective is given by
\begin{equation}
\theta^{*}=\underset{\theta}{\arg \min } ~\mathcal{L}^{c e}\left(\mathcal{D}^{pseudo}_{train} ; \theta\right)~.
\end{equation}

In the \textbf{episodic-training stage}, our goal is to enhance the fast adaptability of $f_\psi$ initialized by $\theta^*$ via training across a variety of few-shot episodes.
In each episode, a subset of $\mathcal{D}_{train}^{pseudo}$ is sampled to construct a few-shot task $\mathcal{T}=(\mathcal{S}_{support}^{pseudo},\mathcal{S}_{query}^{pseudo})$.
The embedding model $f_{\psi}$ makes predictions for the unlabeled query set using the support set~(see Sec.~\ref{sec:testing}).
Formally, the training objective is

\begin{equation}
\psi^*=\underset{\psi}{\arg \min } ~\mathbb{E}_{\mathcal{T}}\left[\mathcal{L}^{ce}\left(
\mathcal{S}_{query}^{pseudo}; \psi, \mathcal{S}_{support}^{pseudo}
\right)\right]~.
\end{equation}
\subsection{Few-shot Testing}\label{sec:testing}
Different from prior work that fine-tunes the embedding model~\cite{Dhillon2020A} or trains a new classifier (\eg, a Fully-Connected layer~\cite{chen2018a}, a Logistic Regression model~\cite{tian2020rethink} or a Support Vector Machine (SVM)~\cite{lee2019meta}) as the base-learner for each task, we keep the $f_\psi$ unchanged and adopt a non-parametric nearest-prototype classifier for few-shot testing.
Given an $N$-way $K$-shot task, let $\mathcal{S}_c\subset\mathcal{S}_{support}$ denote the samples in class $c$, the class prototype $\mathbf{v}_c$ is computed by
\begin{equation}
\mathbf{v}_{c}=\frac{1}{K} \sum_{\mathbf{x}_{i} \in \mathcal{S}_{c}} f_{\psi}\left(\mathbf{x}_{i}\right)~.
\end{equation}
We then predict the probability that a query sample $\mathbf{x}$ belongs to class $c$ using the softmax function on top of the cosine similarity (with temperature $\tau$) between the embedding vector of $\mathbf{x}$ and the class prototype $\mathbf{v}_c$.
Formally,
\begin{equation}
    P(c | \mathbf{x})=\frac{\exp \left(\mathbf{v}_{c}^{T} f_{\psi}(\mathbf{x})/\tau \right)}{\sum^{N}_{k=1} \exp \left( \mathbf{v}_{k}^{T}f_{\psi}(\mathbf{x})/\tau\right)}~.
\end{equation}
\begin{table}[b]
    \centering
    \caption{Dataset Statistics.
    Class numbers are given according to training/validation/testing splits.
    Coarse classes in different splits are non-overlapping.
    Our method largely relieves the annotation burden, as the number of coarse classes is usually far smaller than that of fine classes.
    }
    \setlength{\tabcolsep}{3pt}
    \begin{tabular}{lcccc}\toprule
         \textbf{Dataset}&\textbf{tieredImageNet}&\textbf{FC100}& \textbf{Omniglot} \\\midrule
        Coarse classes & 20/6/8&12/4/4&30/7/13\\
        Fine classes&351/97/160&60/20/20&964/236/423\\
        Images per {\small(fine)} class&$\sim$1282&600&20\\
        Resolution & 84$\times$84  & 32$\times$32 & 28$\times$28\\\bottomrule
         
    \end{tabular}
    \label{tab:dataset}
\end{table}
\section{Experiments}\label{sec:exp}
\subsection{Datasets}
We conduct experiments on three popular benchmark FSL datasets: tieredImageNet~\cite{ren2018meta}, FC100~\cite{tadam2018_66808e32} and Omniglot~\cite{lake2015human}.
These datasets are all \textit{hierarchical} since images are annotated by not only fine labels but also higher-level coarse labels, thus they are suitable as a benchmark for the CGFSC task.
To simulate the situation of inexact supervision, we observe only coarse labels during training.
\textbf{tieredImageNet} is a subset of ILSVRC-12~\cite{russakovsky2015imagenet}, containing 608 classes from 34 super-categories according to the ImageNet~\cite{deng2009imagenet} hierarchy.
The super-categories are divided into 20/6/8 splits for training, validation, and testing respectively, ensuring the training classes to be distinct enough from the testing classes. 
\textbf{FC100} is a derivative of CIFAR-100~\cite{krizhevsky2009learning} with the classes organized in a similar way to tieredImageNet.
\textbf{Omniglot} contains handwritten characters from various alphabets (like Latin and Korean). 
Likewise, we modify the Omniglot by partitioning different alphabets into different splits.
Concretely, we use all alphabets in the \textit{background} split for training, seven alphabets (Manipuri, Atemayar\_Qelisayer, Sylheti, Keble, Gurmukhi, ULOG, Old\_Church\_Slavonic\_(Cyrillic)) for validation, and the other alphabets are left for testing. 
Detailed statistics of the three datasets are summarized in Tab.~\ref{tab:dataset}.

\begin{table}[t]
\caption{
Comparison to fine-to-fine state-of-the-arts and other coarse-to-fine baselines on tieredImageNet.
Average 5-way accuracy (\%) with 95\% confidence intervals.
$^\dag$Upper bound of the coarse-to-fine experiments.
All included methods are with ResNet-12 as the backbone of few-shot learner.
}\label{results:tiered}
\centering
\begin{tabular}{lccccccc}
\toprule
\textbf{Method}&\textbf{1-shot}&\textbf{5-shot}\\\toprule\midrule
\multicolumn{3}{c}{\textit{fine-to-fine~\small{(standard few-shot learning)}}}\\\midrule
Shot-Free~\cite{ravichandran2019few} & 63.52 & 82.59 \\
MetaOptNet~\cite{lee2019meta} & 65.99 $\pm$ 0.72& 81.56 $\pm$ 0.53\\
DSN-MR~\cite{simon2020adaptive} & 67.39 $\pm$ 0.82& 82.85 $\pm$ 0.56\\
RFS-Simple~\cite{tian2020rethink} & 69.74 $\pm$ 0.72& 84.41 $\pm$ 0.55 \\\midrule
Our Baseline~$^\dag$&69.10 $\pm$ 0.79&83.51 $\pm$ 0.56\\\midrule\midrule
\multicolumn{3}{c}{\textit{coarse-to-fine (ours)}}\\\midrule
Baseline w/ Coarse Labels & 57.84 $\pm$ 0.30& 72.21 $\pm$ 0.45\\
Baseline + C2F w/ Pixels& 46.65 $\pm$ 0.73 & 61.04 $\pm$ 0.75\\
VSME w/o C2F~(Ours) &58.49 $\pm$ 0.82&73.14 $\pm$ 0.66 \\
Baseline + C2F w/ VSME~(Ours)&\textbf{60.54 $\pm$ 0.79 }&\textbf{75.22 $\pm$ 0.63}\\
\bottomrule
\end{tabular}

\end{table}
\subsection{Implementation Details}

\heading{Network architecture.}
(1) Regarding the meta-embedder.
We use a standard ResNet-18~\cite{he2016deep} as the backbone, following previous works in unsupervised embedding learning~\cite{wu2018unsupervised,ye2019unsupervised}.
(2) Regarding the base-embedder.
For Omniglot, we apply ConvNet-4 as the few-shot learner following~\cite{vinyals2016matching,finn2017model,zhang2018metagan,li2019gcr}.
ConvNet-4 is composed of 4 modules, each with a 64-filter $3\times3$ convolutional layer followed by batch normalization~\cite{ioffe2015batch} and a ReLU activation. 
A $2\times2$ max-pooling layer is applied after each module.
For tieredImageNet and FC100, we adopt ResNet-12 as commonly-used in~\cite{mishra2018simple,tadam2018_66808e32,lee2019meta,tian2020rethink,ravichandran2019few}.
ResNet-12 contains 4 residual blocks, each having 3 modules, and the first 3 blocks are followed with a $2\times2$ max-pooling.
The number of filters in the blocks is (64,160,320,640).
(3) For all models, the feature maps produced by the backbone are aggregated by a GAP layer to derive low-dimensional embeddings.

\heading{Training details.}
We implement our model using PyTorch~\cite{pytorch}.
We set the embedding dimension $M$ to 128 and the temperature $\tau$ to 0.1.
The number of images per pseudo-fine-class $N_{s}$ for tieredImageNet is 1500, for FC100 is 600, and for Omniglot is 20.
The optimizer is SGD with momentum 0.9, weight decay $5 e^{-4}$, and learning rate decay factor 0.1.
(1) Regarding the meta-embedder.
All experiments are trained for a total of 200 epochs, and the learning rate starts from 0.3 and decays at epoch 120 and 160.
A mini-batch contains 128 samples. 
(2) Regarding the base-embedder.
In the pre-training stage, we train 60 epochs for tieredImageNet, 90 epochs for FC100 and Omniglot.
The learning rate is initialized as 0.05 and decayed twice after the first 30 epochs.
The batch size is 64.
In the episodic training stage, we fix the learning rate to 0.001 and train all models on 2000 pseudo-labeled episodes.

\heading{Evaluation metric.}
To assess the cross-granularity few-shot performance, we evaluate our method on 1000 \textit{fine-labeled} episodes and report the mean accuracy with 95\% confidence intervals. 
Notably, whereas our models learn from only coarse labels, this evaluation protocol is the same as the setting of standard FSL.

\begin{table}[t]
\caption{
Comparison to fine-to-fine state-of-the-arts and other coarse-to-fine baselines on FC100.
Average 5-way accuracy (\%) with 95\% confidence intervals.
$^\dag$Upper bound of the coarse-to-fine experiments.
All included methods are with ResNet-12 as the backbone of the few-shot learner.
}\label{results:fc100}
\centering
\begin{tabular}{lccccccc}
\toprule
\textbf{Method}&\textbf{1-shot}&\textbf{5-shot}\\\toprule\midrule
\multicolumn{3}{c}{\textit{fine-to-fine~\small{(standard few-shot learning)}}}\\\midrule
Prototypical Nets~\cite{snell2017prototypical}& 37.50 $\pm$ 0.60& 52.50 $\pm$ 0.60\\
TADAM~\cite{tadam2018_66808e32}& 40.10 $\pm$ 0.26&  56.10 $\pm$ 0.40\\
MetaOptNet~\cite{lee2019meta}& 41.10 $\pm$ 0.60& 55.50 $\pm$ 0.60\\
RFS-Simple~\cite{tian2020rethink} &42.60 $\pm$ 0.70& 59.10 $\pm$ 0.60 \\\midrule
Our Baseline~$^\dag$&42.39 $\pm$ 0.64&58.91 $\pm$ 0.65\\\midrule\midrule
\multicolumn{3}{c}{\textit{coarse-to-fine (ours)}}\\\midrule
Baseline w/ Coarse Labels & 37.43 $\pm$ 0.66& 52.79 $\pm$ 0.21 \\
Baseline + C2F w/ Pixels&33.99 $\pm$ 0.19&45.31 $\pm$ 0.43\\
VSME w/o C2F~(Ours) &38.22 $\pm$ 0.61& 53.54 $\pm$ 0.60\\
Baseline + C2F w/ VSME~(Ours)&\textbf{39.57 $\pm$ 0.23}&\textbf{54.63 $\pm$ 0.63}\\
\bottomrule
\end{tabular}

\end{table}
\begin{table}[t]
\caption{
Comparison to fine-to-fine state-of-the-arts and other coarse-to-fine baselines on Omniglot.
Average 5-way accuracy (\%) with 95\% confidence intervals.
$^\dag$Upper bound of the coarse-to-fine experiments.
All included methods are with ConvNet-4 as the backbone of the few-shot learner.
}\label{results:omniglot}
\centering
\begin{tabular}{lccccccc}
\toprule
\textbf{Method}&\textbf{1-shot}&\textbf{5-shot}\\\toprule\midrule
\multicolumn{3}{c}{\textit{fine-to-fine~\small{(standard few-shot learning)}}}\\\midrule
Matching Nets~\cite{vinyals2016matching}&98.10& 98.90  \\
MAML~\cite{finn2017model}&98.70 $\pm$ 0.40& 99.90 $\pm$ 0.10 \\
MetaGAN~\cite{zhang2018metagan}&99.67 $\pm$ 0.18& 99.86 $\pm$ 0.11 \\
GCR~\cite{tian2020rethink} &99.72 $\pm$ 0.06& 99.90 $\pm$ 0.10 \\
\midrule
Our Baseline~$^\dag$&98.21 $\pm$ 0.23& 99.48 $\pm$ 0.03\\\midrule\midrule
\multicolumn{3}{c}{\textit{coarse-to-fine (ours)}}\\\midrule
Baseline w/ Coarse Labels & 80.60 $\pm$ 0.26& 93.57 $\pm$ 0.15 \\
Baseline + C2F w/ Pixels&72.87 $\pm$ 0.70&88.53 $\pm$ 0.30\\
VSME w/o C2F~(Ours) & 94.24 $\pm$ 0.23 & 96.12 $\pm$ 0.14\\
Baseline + C2F w/ VSME~(Ours)&\textbf{96.43 $\pm$ 0.32}&\textbf{98.53 $\pm$ 0.04}\\
\bottomrule
\end{tabular}

\end{table}
\subsection{Results on Standard Benchmarks}
\textbf{Baseline models.}
Since the cross-granularity few-shot classification is a new task, we construct a diverse set of baselines to examine the effectiveness of the proposed method.
First, we present several state-of-the-art FSL results for each dataset (in the \textit{fine-to-fine} setting) as a reference.
Second, we also evaluate our few-shot learner (the Base Embedder) under the F2F setting as the upper bound experiments of the C2F counterparts.
Last, we considered the following alternatives as the baselines under the \textit{coarse-to-fine} setting: 
(1)
\textbf{Baseline w/ Coarse Labels.}
Train BE with coarse labels directly (without C2F pseudo-labeling).
(2)
\textbf{Baseline + C2F w/ Pixels.}
Conduct C2F pseudo-labeling with pixel-level features rather than embeddings produced by VSME.
(3)
\textbf{VSME w/o C2F.}
Directly evaluate the VSME with the nearest-prototype classifier rather than using C2F with VSME to generate the pseudo-fine-classes for the downstream few-shot learner to learn from.
(4)
\textbf{Baseline + C2F w/ VSME.}
Our full approach that conducts C2F pseudo-labeling with VSME embeddings and trains a BE as the few-shot learner.
For fairness, we ensure that all experiments on each dataset are carried out with the same backbone architecture of the few-shot learner, and also keep the hyperparameters the same as possible.

\noindent\textbf{Results.}
The results on tieredImageNet, FC100, and Omniglot are shown in Tab.~\ref{results:tiered},~\ref{results:fc100},~\ref{results:omniglot} respectively.
On all datasets, we observe that our approach consistently outperforms the C2F baselines with a large margin.
Especially in the 1-shot setting, which is more challenging compared to the 5-shot setting, we significantly improve the accuracy by 2-16\% across all datasets.
Notably, our results are also comparable to the F2F state-of-the-arts.
More specifically, our method even outperforms \cite{snell2017prototypical} with the same backbone on FC100 by 2.07\% in the 1-shot setting whereas for the 5-shot setting it improves the score by 2.13\%.
In contrast, when C2F pseudo-labeling with pixels, the performance is worse than training with coarse labels, indicating the vital importance of VSME in our method.
We also notice that \textit{VSME w/o C2F} achieves consistent improvements over the other C2F baselines, preceded only by our full model.

\begin{figure}[t]
    \centering
    \includegraphics[width=\linewidth]{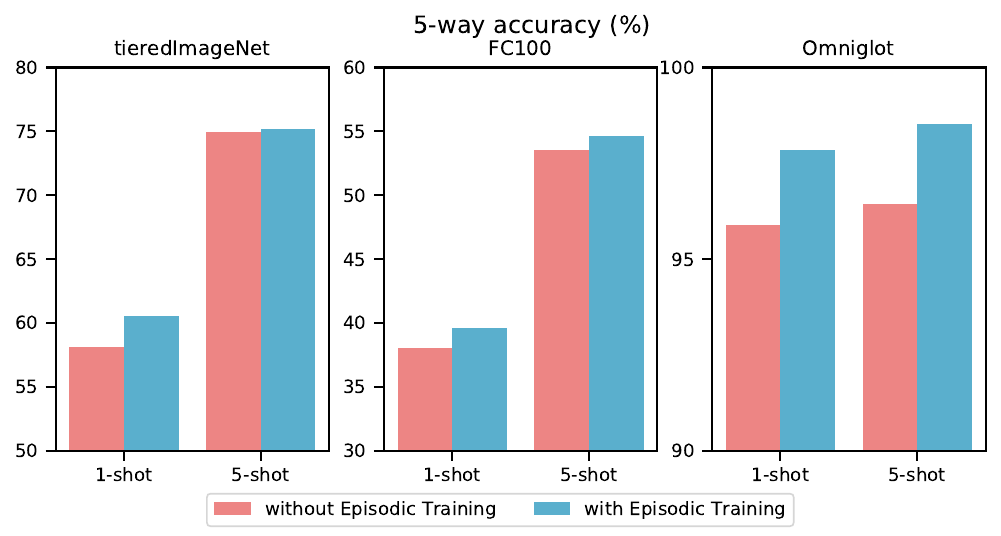}
    \caption{Episodic training brings additional gains.}
    \label{fig:episodic}
\end{figure}

\begin{table}[t]
\centering
\setlength{\tabcolsep}{3.55pt}
\caption{
Effect of the components of VSME on the average 5-way accuracy (\%) on tieredImageNet and Omniglot.
``Dis.'' represents ``discrimination''.
}
\label{tab:ablation}
\begin{tabular}{cccc}
\toprule
 \begin{tabular}{c}\textbf{Visual}\\\textbf{Dis.}\end{tabular}&\begin{tabular}{c}
\textbf{Semantic}\\\textbf{Dis.} \end{tabular}&\begin{tabular}{cc}\multicolumn{2}{c}{\textbf{tieredImageNet}}\\\cmidrule(lr){1-2}\textbf{1-shot}&\textbf{5-shot}\end{tabular}&\begin{tabular}{cc}\multicolumn{2}{c}{\textbf{Omniglot}}\\\cmidrule(lr){1-2}\textbf{1-shot}&\textbf{5-shot}\end{tabular}\\
\midrule
&\checkmark&\multirow{3}{*}{\begin{tabular}{cc}57.10&71.12\\58.06&73.37\\\textbf{60.54}&\textbf{75.22}\\\end{tabular}}&\multirow{3}{*}{\begin{tabular}{cc}76.49&91.19\\81.28&94.26\\\textbf{96.43}&\textbf{98.53}\end{tabular}}\\
\checkmark&&  \\
\checkmark&\checkmark& \\
\bottomrule
\end{tabular}
\end{table}
\subsection{Ablation Study}
In this section, we investigate the contribution of each component.

\heading{Effect of the episodic training.}
Episodic training was widely used in the FSL problems~\cite{vinyals2016matching,snell2017prototypical,li2019gcr}, but recently some works~\cite{tian2020rethink,wang2020cooperative} prefer to merge all episodes together into a single classification task for model training.
In this work, we use it as a warm-up stage to get the model acquainted with tackling the FSL problems after pre-training on the merged task.
As shown in Fig.~\ref{fig:episodic}, in all experiments the episodic training stage consistently improves the 5-way classification accuracy, especially for the 1-shot cases.

\heading{Effect of the components of VSME.}
To show the importance of both the visual- and semantic-discrimination, we evaluate the performance by removing the visual-side item (instance-wise supervision) or the semantic-side item (coarse class-wise supervision) from the full model (Baseline + C2F w/ VSME).
The results are shown in Tab.~\ref{tab:ablation} with the 95\% confidence interval omitted due to space limitation.
We observe that abandoning either of the discrimination would decrease the few-shot classification accuracy on both datasets.
In particular, visual discrimination contributes the most.
We suspect that the reason is that the instance-wise supervision applied for visual discrimination is far stricter than the coarse class-wise supervision for semantic discrimination, as it takes each instance as a separate class.
We also note that these weakened versions still outperform the ``Baseline + C2F w/ Pixels'' baselines in Tab.~\ref{results:tiered} and Tab.~\ref{results:omniglot}.
Another interesting phenomenon is that although when only semantic discrimination is left, the final performance is inferior to ``Baseline w/ Coarse Labels''; it still brings additional gains than applying visual discrimination only.

\begin{table}[t]
    \centering
    \caption{
    Effect of using different few-shot learners on the same pseudo-dataset generated from tieredImageNet.
    }
    \setlength{\tabcolsep}{4.1pt}
    \begin{tabular}{lcccccccccccccc}
         \toprule
         \textbf{Method} &\textbf{Backbone}&\textbf{Setting}&\textbf{1-shot}&\textbf{5-shot}\\
\midrule
\multirow{2}{*}{MAML~\cite{finn2017model}} &\multirow{2}{*}{ConvNet-4} &C2F&47.85 &64.61 \\
&&F2F&51.67&70.30\\
\midrule
\multirow{2}{*}{Prototypical Nets~\cite{snell2017prototypical}} &\multirow{2}{*}{ConvNet-4} &C2F&51.83 &68.71   \\
&&F2F&53.31&72.69\\
\midrule
\multirow{2}{*}{MetaOptNet~\cite{lee2019meta}} &\multirow{2}{*}{ResNet-12}  &C2F& 38.98 &54.20 \\
&&F2F&65.99&81.56\\
\midrule
\multirow{2}{*}{Our Baseline}&\multirow{2}{*}{ResNet-12}&C2F& 60.54&75.22\\
&&F2F&69.10&83.51\\

\bottomrule
    \end{tabular}
    \label{tab:backbone}
\end{table}

\subsection{Further Analysis}
\heading{Effect of the few-shot learner.}
Since we generate pseudo-fine-labeled datasets as the intermediate results, our framework can be integrated with other FSL algorithms.
Here, we choose~\cite{finn2017model,snell2017prototypical,lee2019meta} to study the effect of the few-shot learner.
We first conduct C2F pseudo-labeling on tieredImageNet to generate a pseudo-dataset and then train different few-shot learners on it.
As reported in Tab.~\ref{tab:backbone}, the methods using shallower backbone (ConvNet-4) suffer less performance drop facing the cross-granularity transfer challenge, but meanwhile, show lower C2F accuracy.
The performance of~\cite{lee2019meta} dramatically drops because it reinitializes a maximum-margin classifier (SVM) for each task, easily misled by the inaccurate pseudo-labels.

\heading{Robustness analysis.}
As described in Alg.~\ref{alg:1}, the C2F pseudo-labeling algorithm randomly selects from the unlabeled images as the root to establish a new pseudo-fine-class, involving a series of random sampling.
The number of samples per pseudo-fine-class $N_s$ is another key parameter in the C2F algorithm.
To verify the robustness of our method, we conduct experiments across 3 different $N_s$, and for each $N_s$ we repeat 3 times with different random seeds. 
The results of 5-way 1-shot, 5-shot, and 10-shot FSL tasks on the FC100 dataset are presented in Tab.~\ref{tab:robustness}.
It is evident that our method is fairly stable and robust to these critical hyperparameters.

\heading{Effect of the embedding dimension \& the number of shots.}
According to Tab.~\ref{tab:robustness}, increasing both the embedding dimension $M$ or the number of shots $K$ tends to improve the CGFSC accuracy.

\begin{table}[t]
    \centering
    \caption{
    Average 5-way accuracy (\%) with 95\% confidence intervals on FC100.
    $\mathbf{M}$: The embedding dimension of VSME.
    $\mathbf{N_s}$: Number of samples per pseudo-fine-class in the C2F process.
    $\mathbf{seed}$: Random seed in the C2F process.
    }
    \begin{tabular}{ccccccc}\toprule
         $\mathbf{M}$&$\mathbf{N_s}$ & $\mathbf{seed}$ &\textbf{1-shot}&\textbf{5-shot}&\textbf{10-shot}\\\midrule
         \multirow{9}{*}{128}& \multirow{3}{*}{550}&0&39.02 $\pm$ 0.23&54.19 $\pm$ 0.63&59.27 $\pm$ 0.62\\
         &&1& 39.03 $\pm$ 0.36&54.25 $\pm$ 0.64&59.15 $\pm$ 0.36\\
         &&2&39.04 $\pm$ 0.23&53.96 $\pm$ 0.25& 59.10 $\pm$ 0.21\\\cline{2-6}
         &\multirow{3}{*}{600}&0&39.53 $\pm$ 0.35&54.43 $\pm$ 0.36&59.75 $\pm$ 0.36\\
         &&1&39.57 $\pm$ 0.23&54.33 $\pm$ 0.25&59.65 $\pm$ 0.27\\
         &&2&39.31 $\pm$ 0.79 &54.63 $\pm$ 0.63&59.99 $\pm$ 0.58\\\cline{2-6}
         &\multirow{3}{*}{650}&0&39.04 $\pm$ 0.35&53.35 $\pm$ 0.20&58.39 $\pm$ 0.63\\
         &&1&38.93 $\pm$ 0.23&53.50 $\pm$ 0.65&58.26 $\pm$ 0.63\\
         &&2&38.99 $\pm$ 0.31&53.34 $\pm$ 0.26&58.25 $\pm$ 0.36
         \\\cmidrule{1-6}
         \multirow{9}{*}{256}& \multirow{3}{*}{550}&0&39.61 $\pm$ 0.27&54.73 $\pm$ 0.61&59.76 $\pm$ 0.61\\
         &&1&39.61 $\pm$ 0.30&55.05 $\pm$ 0.63&59.96 $\pm$ 0.61\\
         &&2&39.62 $\pm$ 0.23&54.65 $\pm$ 0.45&59.72 $\pm$ 0.35\\\cline{2-6}
         &\multirow{3}{*}{600}&0&39.53 $\pm$ 0.66&54.71 $\pm$ 0.62&59.76 $\pm$ 0.27\\
         &&1&39.43 $\pm$ 0.32&55.00 $\pm$ 0.62&59.82 $\pm$ 0.35\\
         &&2&39.61 $\pm$ 0.65&54.58 $\pm$ 0.36&60.06 $\pm$ 0.57\\\cline{2-6}
         &\multirow{3}{*}{650}&0&38.98 $\pm$ 0.66&53.80 $\pm$ 0.24&58.86 $\pm$ 0.35\\
         &&1&38.95 $\pm$ 0.36&53.97 $\pm$ 0.63&58.90 $\pm$ 0.27\\
         &&2&38.96 $\pm$ 0.66&53.79 $\pm$ 0.31&58.88 $\pm$ 0.21
         \\\bottomrule
    \end{tabular}
    \label{tab:robustness}
\end{table}
\section{Conclusion}
In this work, we propose an inexactly-supervised meta-learning system, motivated to tackle FSL problems with only coarse labels for reducing the annotation burden. 
To address this cross-granularity challenge, we propose to meta-learn a Meta-Embedder to synthesize a class distribution that has analogous granularity to the true distribution.
We jointly optimize the visual- and semantic-discrimination of the image embeddings and introduce a C2F pseudo-labeling algorithm to generate pseudo-fine-classes via greedy-based similarity matching on the learned embeddings.
We also develop a two-stage strategy to train the Base-Embedder on the pseudo-dataset for FSL.
Experimental results and ablation studies have indicated that our framework achieves substantial improvements over the baselines and is comparable to the fine-grained upper bounds. 

\begin{acks}
This work was funded by National Natural Science Foundation of China (NSFC, Grant No. 61771303), Science and Technology Commission of Shanghai Municipality (STCSM, Grant Nos. 19DZ1209303, 20DZ1200203, 2021SHZDZX0102), and SJTUYitu/Thinkforce Joint Laboratory for Visual Computing and Application.
\end{acks}

\bibliographystyle{ACM-Reference-Format}
\bibliography{main}

\end{document}